\pdfoutput=1

\documentclass[11pt]{article}

\usepackage[preprint]{acl}

\usepackage{times}
\usepackage{latexsym}

\usepackage[T1]{fontenc}

\usepackage[utf8]{inputenc}

\usepackage{microtype}

\usepackage{inconsolata}

\usepackage{graphicx}
\usepackage{enumitem}

\usepackage{blindtext}
\usepackage{breakcites}
\title{\dataname{}:
A Large-Scale Multi-Dimensional Predictions Dataset\\ Towards Meaningful LLM Evaluation}

\author{
  \textbf{Eliya Habba\textsuperscript{1}},
  \textbf{Ofir Arviv\textsuperscript{2}},
  \textbf{Itay Itzhak\textsuperscript{1,4}},
  \textbf{Yotam Perlitz\textsuperscript{2}},
\\
  \textbf{Elron Bandel\textsuperscript{2}},
  \textbf{Leshem Choshen\textsuperscript{2,3}},
  \textbf{Michal Shmueli-Scheuer\textsuperscript{2}},
  \textbf{Gabriel Stanovsky\textsuperscript{1,5}}
\\
\\
  \textsuperscript{1}The Hebrew University of Jerusalem,
  \textsuperscript{2}IBM Research AI,
  \textsuperscript{3}MIT,\\
  \textsuperscript{4}Technion - Israel Institute of Technology,
  \textsuperscript{5}Allen Institute for AI
\\
  {
\href{mailto:eliya.habba@mail.huji.ac.il}{eliya.habba@mail.huji.ac.il}  }
}

\usepackage{listings}
\usepackage{xcolor}
\usepackage{graphicx}
\usepackage{pgf}
\usepackage[normalem]{ulem}
\usepackage{makecell}
\usepackage{multirow}
\useunder{\uline}{\ul}{}
\usepackage{ulem} 
\usepackage[export]{adjustbox}
\usepackage{xcolor}  
\usepackage{booktabs}  
\usepackage{tabularx}
\usepackage{booktabs}
\usepackage{tabularx}
\usepackage{caption}
\usepackage{geometry}
\usepackage{amsmath}
\usepackage{graphicx}
\usepackage{svg}
\usepackage{subcaption} 
\usepackage{calc}
\usepackage{setspace}  
\usepackage{enumitem}  
\usepackage{listings}  
\usepackage{bbm}
\usepackage{float} 
\usepackage{arydshln}

\svgpath{{figures/}}
\newcommand{\olmo}[0]{OLMo}
\newcommand{\dataname}[0]{\textsc{Dove}}
\newcommand{\datasetsize}{250M}
\newcommand{\explicitdataname}[0]{Dataset Of Variation Evaluation}

\begin{document}
\maketitle
\begin{abstract}
\label{sec:abstract}
 Recent work found that LLMs are sensitive to a wide range of arbitrary prompt dimensions, including the type of delimiters, answer enumerators, instruction wording, and more. This throws into question popular single-prompt evaluation practices.
We present \dataname{}~(\explicitdataname{})
a large-scale dataset containing prompt perturbations of various evaluation benchmarks. In contrast to previous work, we examine LLM sensitivity from an \emph{holistic} perspective, and assess the joint effects of perturbations along various dimensions, resulting in thousands of perturbations per instance. We evaluate several model families against \dataname{}, leading to several findings, including efficient methods for choosing well-performing prompts, observing that few-shot examples reduce sensitivity, and identifying instances which are inherently hard across all perturbations.
\dataname{} consists of more than \datasetsize{}
prompt perturbations and model outputs, which we make publicly available to spur a community-wide effort toward meaningful, robust, and efficient evaluation.





\vspace{0.5em} 
\textbf{Browse the data, contribute, and more at:}
\newline\noindent\url{https://slab-nlp.github.io/DOVE}



\end{abstract}

\section{Introduction}
\label{sec:introduction}


\begin{figure}[tb!]
    \centering
    \includegraphics[width=\columnwidth]{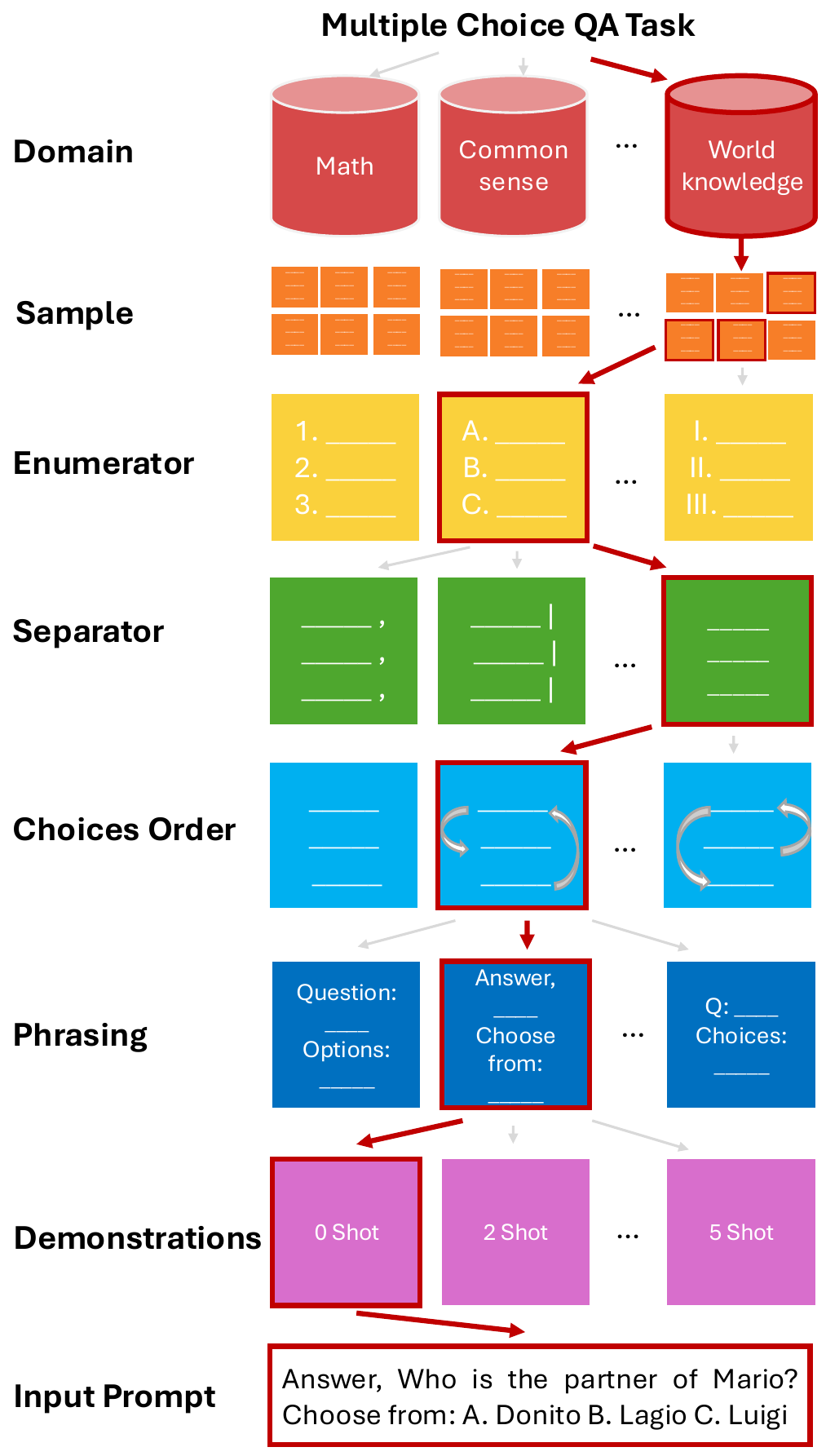}
    \caption{\textbf{Building \dataname{}.} To holistically explore LLM sensitivity, we sample prompts as a walk in the space of various \emph{prompt dimensions} (rows, above).}
    \label{fig:rubiks-cube}
\end{figure}

Recent years have seen an explosion of LLMs applied in few- or zero-shot settings, where natural language is used for both input and output. Although this free-text format lends itself to various applications, the flexibility in task formulation also leads to large variation in performance. 

LLM performance was shown to change drastically based on slight perturbations in arbitrary prompt dimensions, including the number of white spaces~\cite{sclar2024quantifyinglanguagemodelssensitivity}, 
answer enumerators and ordering~\cite{alzahrani-etal-2024-benchmarks,pezeshkpour-hruschka-2024-large},
few-shot demonstrations~\cite{lu-etal-2022-fantastically}, and more~\cite{leidinger-etal-2023-language,voronov-etal-2024-mind}.
This sensitivity presents a challenge to meaningful evaluation, exacerbated by the rising cost of inference, which bars large-scale evaluation studies, especially for research groups with small to medium budgets~\cite{perlitz-etal-2024-efficient}.

Such concurrent findings throw into question  
the generalizbility of many of the recent evaluation benchmarks, which tend to rely on one arbitrary prompt~\cite{mizrahi-etal-2024-state}. 
We argue that this constitutes a crisis in evaluation which should be a \emph{community-wide concern}, standing in the way of a better scientific understanding of LLMs, indicating where they excel and where they lack, especially as they are being increasingly deployed in real-world applications~\cite{10433480}.

Our main contribution in this work is the introduction of \dataname{}, a publicly available large-scale dataset consisting of \datasetsize{} \emph{model predictions}, which facilitates and democratizes the systematic study of LLM sensitivity and the development of meaningful evaluation protocols.

Starting from popular multiple-choice benchmarks, such as MMLU~\cite{hendryckstest2021}, ARC~\cite{DBLP:journals/corr/abs-1803-05457}, or Race~\cite{lai-etal-2017-race},
we go beyond common evaluation protocols and collect LLM predictions on a \emph{wide range of prompt perturbations},
resulting in thousands of samples per single instance from the original benchmark. 
For each such instance, \dataname{} records the full LLM response along with the model's log probabilities and an automatic binary score. 

We analyze the performance of various LLMs on \dataname{}  and find that the problems observed at smaller scales persist at this large scale. We find that along various dimensions (prompt phrasing, formatting, and more), performance can vary by more than 10\% absolute difference, while model ranking also varies based on these arbitrary choices. These make \dataname{} a valuable testbed for exploring evaluation and sensitivity at scale. 

To demonstrate the kind of analysis permitted by \dataname{}, we use it to make three novel observations on prompt sensitivity in LLMs, which benefit downstream application and provide a more meaningful evaluation. First, we observe that prompt-tuning the entire prompt is subpar compared to independent dimension-wise tuning; second, we find that adding few-shot demonstrations consistently reduces sensitivity, though it is far from solving the problem; and third, \dataname{} can be used to find consistently hard instances, which stump models regardless of any prompt selection, thus delineating the real limits to their capabilities.



By making \dataname{} publicly and openly available, we hope to enable and spur research into meaningful, generalizable, and efficient LLM evaluation, which will help to understand their strengths and limitations. Toward that goal, we plan to make \dataname{} a collaborative and growing resource and encourage the contribution of data from more diverse domains, applications, and languages.

\section{Definitions: Prompt Sensitivity}
\label{sec:background}


In this section, we establish terminology, definitions, and metrics for formally quantifying the phenomenon of prompt sensitivity. In this work, we choose to focus on multiple-choice questions to allow for a relatively easy evaluation of model outputs compared to text generation tasks, such as summarization or translation, where the space of correct predictions is vast, and may be considered in future work.

\paragraph{Intent-preserving prompts.} Following \citet{Chatterjee2024POSIXAP}, two prompts $p_1, p_2$ are considered \emph{intent-preserving} if they are designed to convey the same underlying meaning, despite differences in phrasing or structure. For example, the two following prompts are considered intent preserving $p_1$ = \emph{``Who is the partner of Mario? Choose from: A. Donito B. Lagio C. Luigi''}, $p_2$ = \emph{``Answer the following question: Who is the partner of Mario? A. Donito B. Lagio C. Luigi''.}

\paragraph{Prompt dimensions and linearization.} We  categorize the differences between intent-preserving prompts along different dimensions, where each dimension $D$ is a set of possible values such that any value from $d \in D$ preserves the intent of the prompt. For example, the \emph{enumerator} dimension may contain values such as \{roman, numerals\}. Like enumerators, prompt dimensions may be discrete or continuous, e.g., instruction paraphrase. Furthermore, we define \emph{prompt linearization}:
\begin{equation}
T(x, d_1, \ldots, d_n) \mapsto p 
\end{equation}
Where $x$ is an underlying question, e.g., \emph{``who is Mario's partner?''}, 
$d_1 \in D_1, \ldots, d_n \in D_n$ are choices made along $n$ prompt dimensions, and $T$ is their deterministic linearization to a prompt $p$, which can be given as input to an LLM.

\paragraph{Prompt sensitivity.}  
measures the degree to which the performance of an LLM $M$ deviates between intent-preserving prompts.
Ideally, the performance of an LLM $M$ should be invariable to different choices along intent-preserving prompt dimensions. Formally, to measure prompt sensitivity on multiple-choice questions, we define a model $M$'s accuracy along different dimension choices $d_1, \ldots, d_n$ in the following manner:
\begin{equation}
\label{eq:acc}
    \begin{aligned}
    A&cc(M, Dom, d_1, \ldots, d_n) = \\
    & \frac{\sum\limits_{(x_i, y_i) \in Dom} \mathbbm{1}(M(T(x_i, d_1, \ldots, d_n)) = y_i)}{|Dom|}
\end{aligned}
\end{equation}

Where $Dom$ is a dataset consisting of labeled tuples $(x_i, y_i)$ in a certain domain, for example (\emph{who is Mario's partner?, Luigi}). Intuitively, $Acc$ measures the accuracy of $M$ on $Dom$ according to a specific set of choices for the different prompt dimensions. Consequently, we measure \emph{prompt sensitivity} as the difference in accuracy for different dimensions using various statistical measures.



\section{\dataname{}: A Large-Scale Multi-Dimensional Dataset of LLM-Generated Responses Towards Meaningful LLM Evaluation}
\label{sec:dataset_creation}

In this section we introduce \dataname{}, a large-scale corpus of model predictions along multiple dimensions. 

\begin{figure}[tb!]
    \centering
  \includegraphics[width=\linewidth]{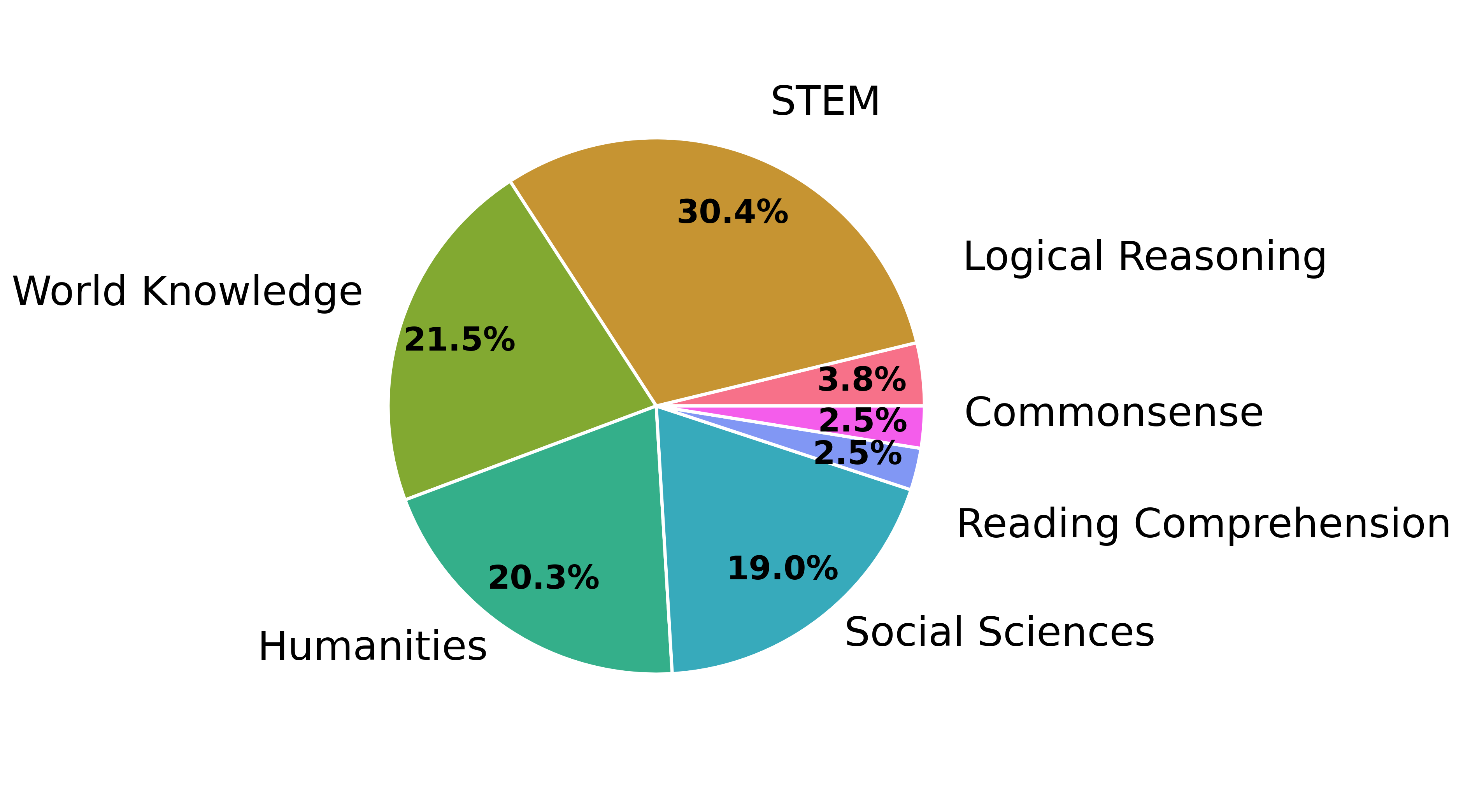}
    \caption{\dataname{} requires a diverse set of skills.} 
\label{fig:skills}
\end{figure}

As shown in Figure~\ref{fig:rubiks-cube}, the building blocks of \dataname{} are instances from existing popular datasets.
For each instance, we create a wide range of intent-preserving prompts, by varying the instances along five dimensions (enumerator, separator, choices order, phrasing, and demonstrations).
\begin{figure*}[tb!]
    \centering
    \includegraphics[width=\linewidth]{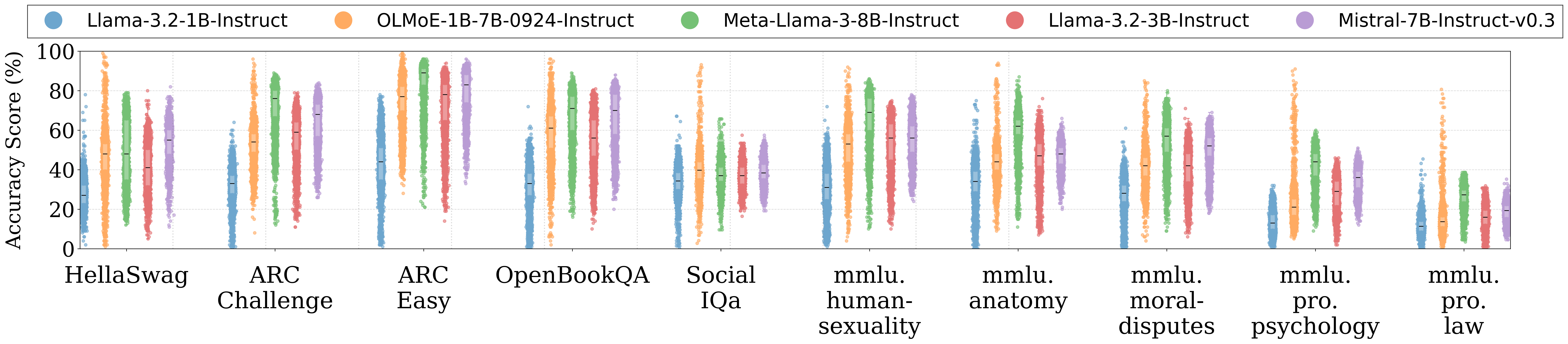}

    \caption{\textbf{Performance variations across evaluation datasets.} Each datapoint represents the accuracy of one model calculated across 100 instances.
    Vertical scatter plots
    illustrate the variance within each dataset and each model. Model performance varies substantially, 
    indicating persistent prompt sensitivity prompts at large scales.
    }
    
    \label{fig:model_performance}
\end{figure*}


            
 

\begin{table}[tb!]
    \centering
    \renewcommand{\arraystretch}{1.2} 
    \setlength{\tabcolsep}{5pt} 
    \resizebox{\columnwidth}{!}{ 
        \begin{tabular}{l m{5cm} m{1.2cm}} 
    \toprule
    \textbf{Dimension} & \multicolumn{1}{l}{\textbf{Examples}} & \multicolumn{1}{r}{\textbf{\# of Values}} \\

    \midrule
    \textbf{Enumerator} & Roman, Numerals & 6   \\\hline
    \textbf{Separator} & ;, | & 7  \\\hline
    \textbf{Choices Order} & original, correct first & 6  \\\hline
    \textbf{Phrasing} & \textit{The following are multiple-choice questions about \textbraceleft topic \textbraceright. \textbraceleft question \textbraceright \textbraceleft choices \textbraceright Answer:} & 13   \\\hline
    \textbf{Demonstrations} & Zero-shot, Five-shot & 2  \\

    \bottomrule
    \end{tabular}
    }
    \caption{The different intent-preserving prompt dimensions in \dataname{}, along with example values, and overall number of values per dimension. The total number of perturbations per sample is the Cartesian product of all values, resulting in over 6.5K perturbations per sample.}
    \label{table:prompt_dimensions_table2}
\end{table}

Below we discuss the different  dimensions, which are also summarized in Table~\ref{table:prompt_dimensions_table2}. 
We choose these dimensions based on a survey of recent studies on LLM sensitivity, yet we do not claim that this forms an exhaustive list of factors affecting LLM performance. Future work can expand this with additional dimensions to explore their effect. 


\paragraph{Domains.}
We cover a wide range of data sources, spanning 78 different data sets from MMLU~\cite{hendryckstest2021}, MMLU Pro, ARC~\cite{DBLP:journals/corr/abs-1803-05457}, HellaSwag~\cite{zellers2019hellaswag}, OpenBookQA~\cite{OpenBookQA2018}, Social IQa~\cite{sap-etal-2019-social}, and RACE~\cite{lai-etal-2017-race}.
From each of these, we take 100 instances chosen at random, resulting in 7,800 base instances, which we extend with different perturbations in subsequent steps.
Figure~\ref{fig:skills} shows that solving these samples requires a wide range of skills.

\paragraph{Answer enumerators, choice separators and orderings.} 
Recent work has noticed that very subtle changes in the prompt can lead to significant changes in both absolute as well as relative model performance. 
These include answer enumerators, e.g., roman versus numeral options, choice separators, e.g., new line versus commas, and the order in which the options are presented, e.g., the position of the correct answer~\cite{alzahrani-etal-2024-benchmarks,pezeshkpour-hruschka-2024-large,zhou-etal-2024-revisiting,gupta2024changinganswerorderdecrease}. 
All options are summarized in Table~\ref{tab:prompt-formatting-dimensions} in the Appendix.


\paragraph{Instruction phrasing.}
Variations in the way instructions are written can significantly influence model behavior~\cite{mizrahi-etal-2024-state,Chatterjee2024POSIXAP}. 
To systematically explore this effect, we wrote and verified 13 distinct instruction templates for each of our datasets, drawing inspiration from the format used in established benchmarks like MMLU \cite{hendryckstest2021} and HELM \cite{liang2023holistic}, as well as paraphrases from \citet{zhuo2024prosaassessingunderstandingprompt} and \citet{mizrahi-etal-2024-state}. See Appendix~\ref{sec:instruction_phrasing_options} for a complete listing of paraphrased instructions.

\paragraph{Demonstrations.}  
We vary the number of few-shot demonstrations, chosen randomly from the training set of each dataset, based on previous work which found this to be a factor affecting model performance~\cite{zhao2021calibrateuseimprovingfewshot,lu-etal-2022-fantastically,kumar-talukdar-2021-reordering,reif-schwartz-2024-beyond}.

\paragraph{Additional metadata.}
Table~\ref{tab:metada} shows additional instance-level details available in \dataname{} to allow future research into their effect, such as the input and output log probabilities assigned by the model.

\begin{table}[tb!]
\centering
\renewcommand{\arraystretch}{1.2}
\small
\begin{tabular}{p{2.6cm} p{4.4cm}}
\toprule
\textbf{Field} & \textbf{Description} \\ \midrule
\textbf{Hyperparameters} & Temperature, top-p \\ \hline
\textbf{Tokens logprobs} & Model's log probability of prompt tokens \\ \hline
\textbf{Few-shots} & Example question-answer pairs \\ \hline
\textbf{Response} & Model's full response to the prompt \\\hline
\textbf{Tokens logprobs} & Model's log probabilities for generated tokens \\ \hline
\textbf{Ground truth} & The correct answer for the given instance \\\hline
\textbf{Evaluation method} & Name of method used to evaluate the model's response \\\hline
\textbf{Score} & Automatic evaluation score \\ \bottomrule
\end{tabular}
\caption{\textbf{Additional metadata}. Instance-level details available in \dataname{} to allow future research into their effect, such as the input and output log probabilities assigned by the model.
}
\label{tab:metada}
\end{table}


\section{Evaluation}
In this section, we evaluate various models against \dataname{}, finding that they all exhibit prompt sensitivity at large scale, also when controlling for most of our tested dimensions.
\label{sec:evaluation}
\subsection{Experimental Setup}
We evaluate the following model families against \dataname{}:
Llama (1B, 3B, 8B)~\cite{dubey2024llama}, \olmo{} (7B)~\cite{muennighoff2024olmoeopenmixtureofexpertslanguage}, and Mistral (7B)~\cite{jiang2023mistral7b}. We focus on \emph{open-weight} LLMs which we can run locally for two main reasons. 
First and foremost, API-based chatbots (such as ChatGPT or Claude) alter the prompt in undisclosed ways, 
for example, to try to ensure that it is safe, or to improve performance~\cite{rao2023tricking}, which may interfere with our findings in a non-trivial manner. 


Second, running closed models in such a large scale (60M instances per model) incurs infeasible costs, which do not pay back to the community.  However, we note that such sensitivity  was observed in closed models~\cite{mizrahi-etal-2024-state} as well as large open models~\cite{zhou2024revisiting,alzahrani2024benchmarks,gupta2024changing}, suggesting that these phenomena are not artifacts of model size limitations, and we encourage future work to test them on \dataname{}.

We generate \dataname{} using vLLM \cite{kwon2023efficientmemorymanagementlarge} on a cluster of NVIDIA A100 80GB GPUs. In total, dataset creation requires approximately 5,000 GPU hours. For instance, the Mistral-7B model requires 1,189 GPU hours, while other models range from 754 to 1,341 GPU hours each. Overall, creating \dataname{} on cloud services, such as AWS, costs upwards of \$25K, highlighting the high costs of such large scale evaluations.

We extend and use the Unitxt framework~\cite{bandel2024unitxt} to generate and evaluate multiple prompt variations in multiple datasets.

\paragraph{Evaluation metric.}
To evaluate model outputs we use semantic similarity matching \cite{mitkov2009semantic,obot2023grading}. For each response, we identify the answer option with highest semantic similarity to the model's output and consider the prediction correct if it matches the ground truth.

\begin{figure*}[tb!]
    \centering
  \includegraphics[width=\linewidth]{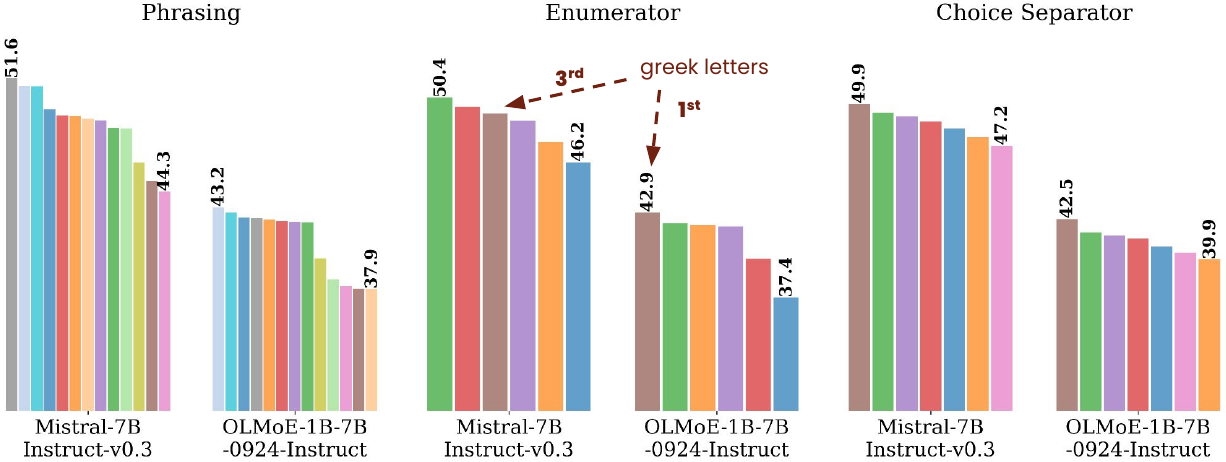}
  \caption{\textbf{Accuracy marginalization for different dimensions.} Variations along each of the dimensions in \dataname{} lead to prompt sensitivity, even when controlling for all other dimensions.}
\label{fig:marginalization}
\end{figure*}

\subsection{Results: Prompt Sensitivity Persists in Large-Scale Data }

Figure~\ref{fig:model_performance} depicts model performance on several domains as a distribution across intent-preserving prompts, while similar trends were observed across all other domains~(see Appendix~\ref{sec:performance_analysis_all_domains}). 
For instance, \olmo{}'s performance on HellaSwag ranges from 1\% to 99\% based on the prompt.
These findings suggest that the dimensions we explore in \dataname{} indeed play a role in the performance of all LLMs.

To better understand these results, 
we marginalize each dimension by averaging its performance across all other dimensions. 
Formally, without loss of generality for each value $d_1 \in D_1$ (for example, the choice of roman numerals), we compute a marginalized accuracy score $Acc_{d_1}$:

{\footnotesize
\begin{equation}
Acc_{d_1}(M, Dom) = \sum\limits_{\substack{d_2 \in D_2 \\ \vdots \\ d_n \in D_n}} \frac{Acc(M, Dom, d_1, \ldots, d_n)}{|D_2|\cdot\ldots\cdot|D_n|}
\end{equation}
}

Where $D_1, \ldots, D_n$ are the different dimensions, and $Acc(\cdot)$ is according to Equation~\ref{eq:acc}.

The results, depicted in Figure~\ref{fig:marginalization} show that variation along each individual dimension changes results substantially. 
For instance, for Mistral, different paraphrases lead to an 8\% difference in accuracy. Beyond absolute performance differences, we also observe varying preferences across models to different prompt variations. For example, OLMoE performs best with greek numerals, achieving the highest average accuracy across the dataset with this choice. On the other hand, Mistral rank greek numerals only as the third best option, performing less than both capital and lateen numerals. This discrepancy underscores that models demonstrate distinct prompt preferences.

\begin{figure}[tb!]
    \centering
  \includegraphics[width=\linewidth]{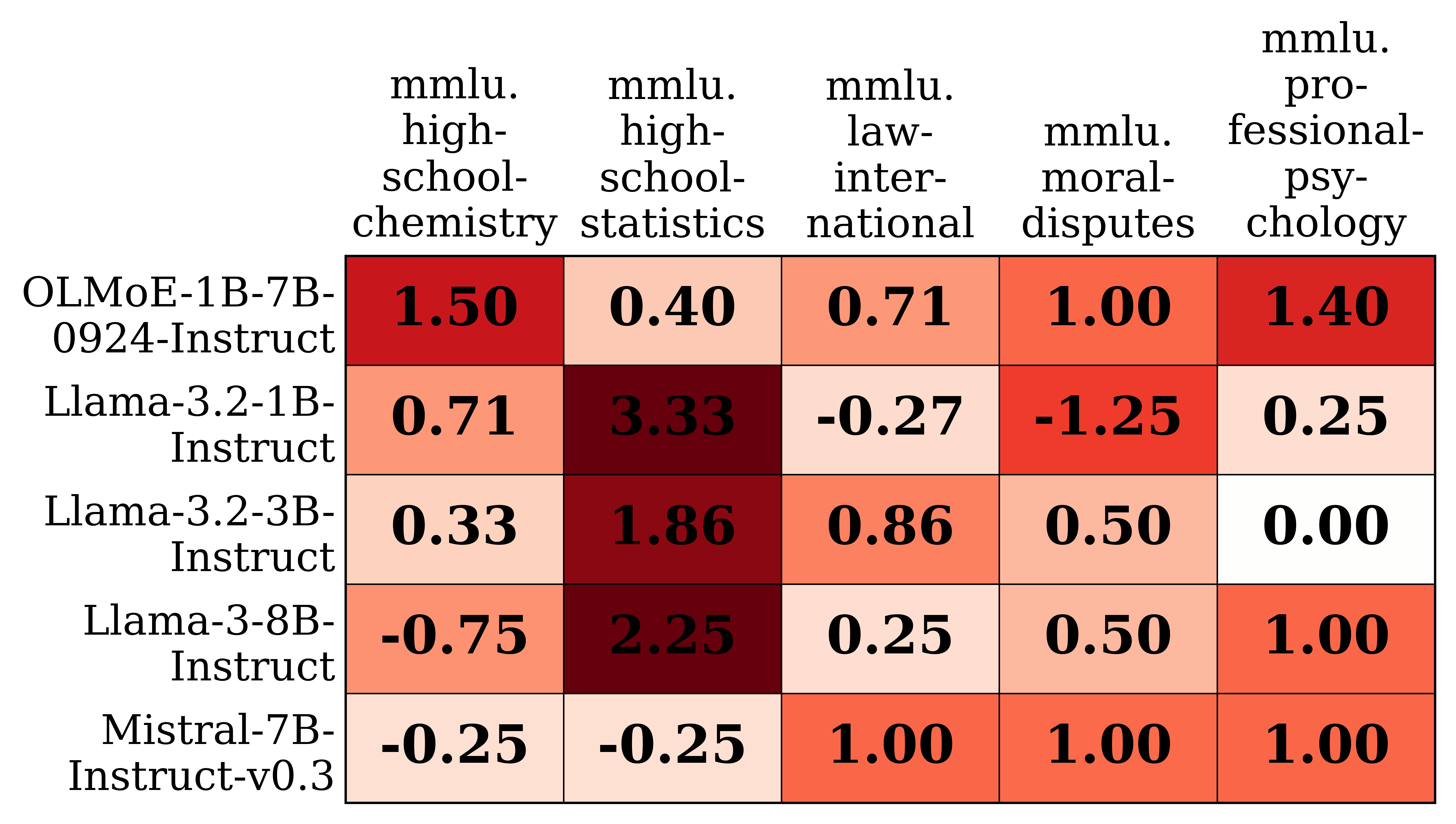}
  \caption{\textbf{Substantial performance differences across prompt perturbation.} The number of standard deviations by which model performance on original instructions deviates from average across few-shot prompts. Dark cells show substantial divergence.}
\label{fig:divergence_plot}
\end{figure}

\paragraph{Statistical significance.}
Following ~\cite{mizrahi-etal-2024-state}, we quantified performance variance by calculating divergence scores, defined as the number of standard deviations by which performance using the original prompt deviates from the mean performance across all prompts. 
Figure~\ref{fig:divergence_plot} shows significant divergence in randomly sampled domains from the MMLU~\cite{hendryckstest2021}, where divergence is defined as exceeding one standard deviation \cite{kazmier2003business}. For Instance, Mistral's performance with original prompts exceeds its mean performance by more than one standard deviation in 35 of 57 domain tasks (complete results can be found in Figure~\ref{fig:appendix_divergence_plot} in Appendix~\ref{sec:divergence_across_all_domains})

\label{fig:distribution_template}

\section{Analysis}
\label{sec:analysis}
So far, we made use of \dataname{} to quantify the effect of prompt sensitivity in large scale, finding that each of the individual prompt dimensions further contributes to this sensitivity.
In this section, we discuss three observations that stem from this large-scale analysis and have practical implications for downstream applications and for more generalizable and meaningful evaluation. 


\begin{figure*}[tb!]
    \centering
    \includegraphics[width=\linewidth]{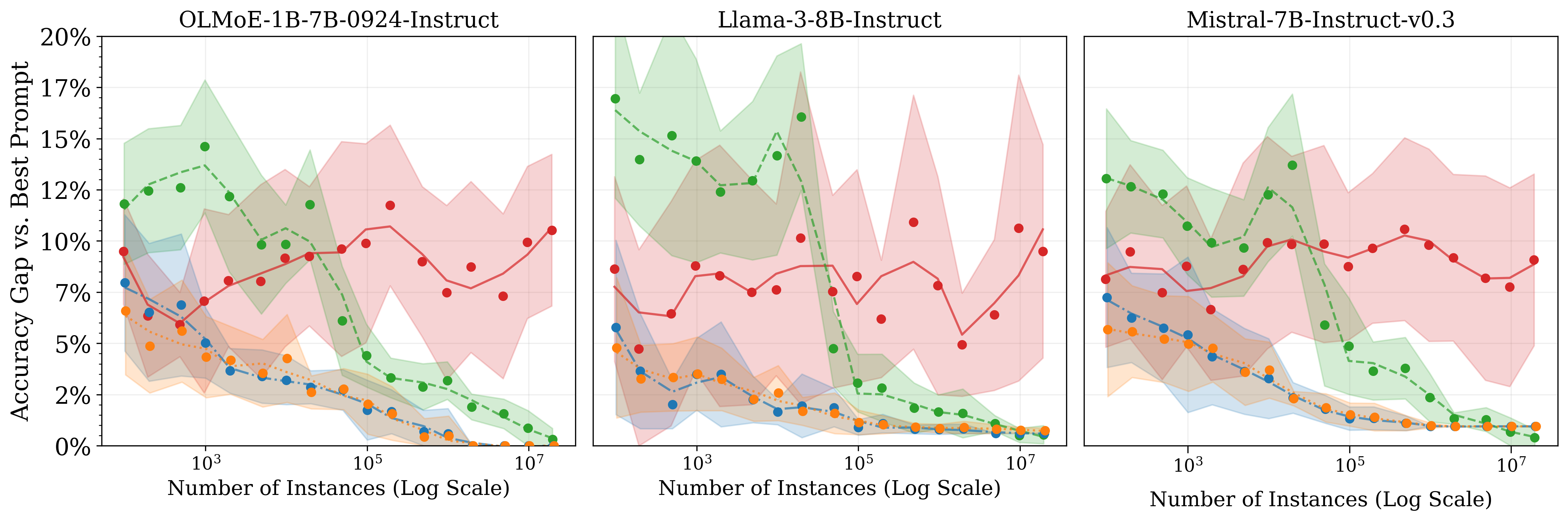}
    
    \vspace{0.cm} 
    
    \includegraphics[width=1\linewidth]{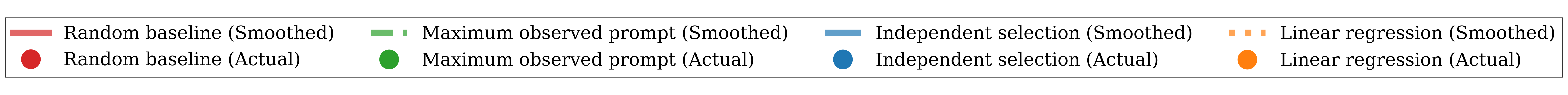}

    \vspace{0.cm} 

    \caption{\textbf{Efficient prompt selection approaches can improve perfromance.} Performance gap from the ground truth prompt (y-axis) versus sample count (x-axis) for LLMs and selection methods. Results demonstrate that efficient prompt selection methods can improve  performance with relatively small sample sizes, outperforming random selection.}
    \label{fig:efficient}
\end{figure*}
\begin{figure}[tb!]
    \centering
    \includegraphics[width=\linewidth]{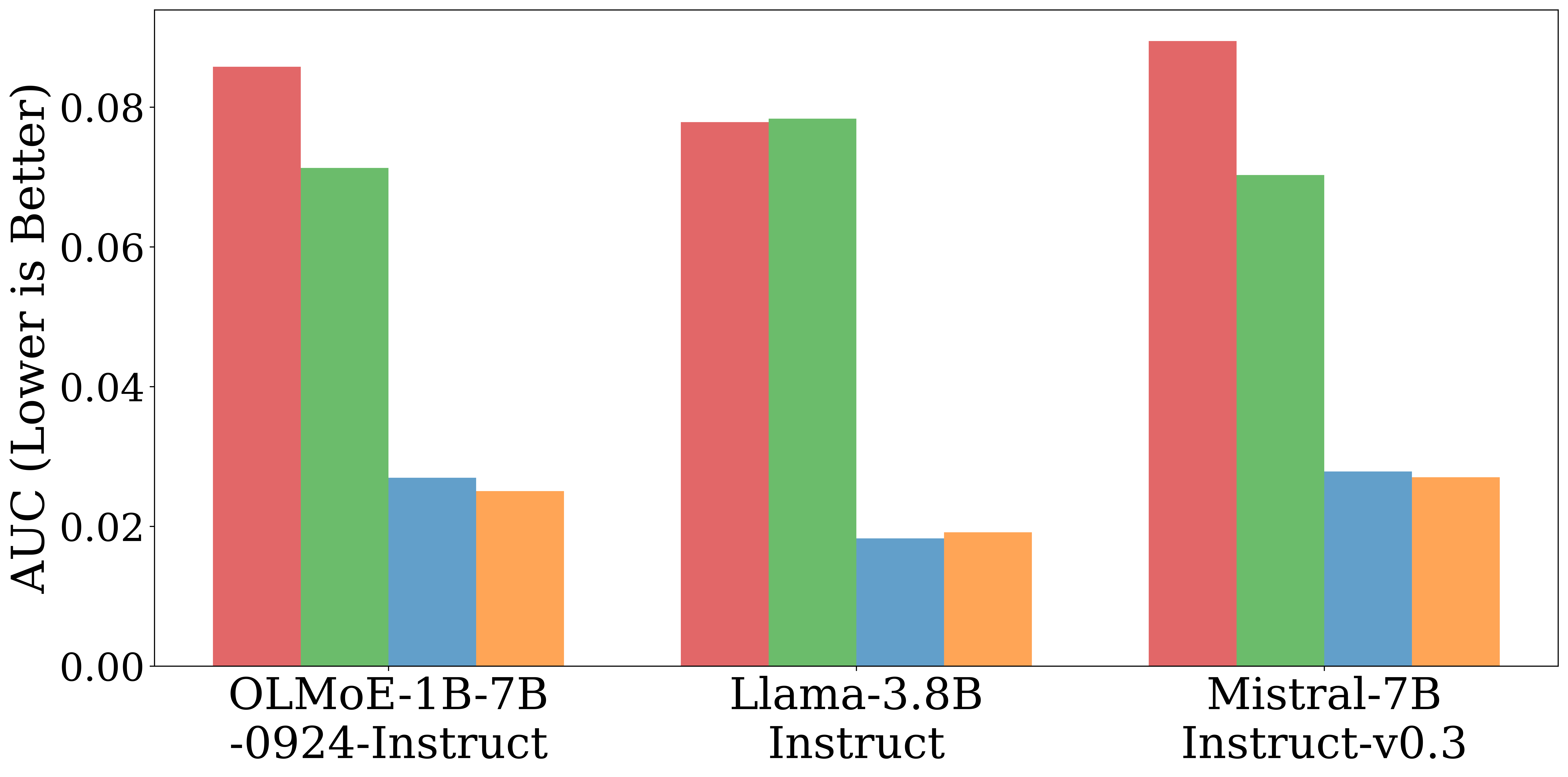}

    \vspace{0.cm} 
    
    \includegraphics[width=1\linewidth]{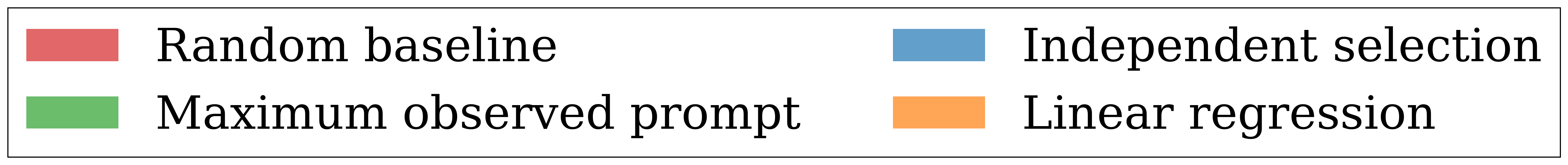}

    \vspace{0.cm} 
    
    \caption{\textbf{Prompt selection methods outperform random and best observed baseline.} 
    AUC comparison of prompt selection methods across different LLMs. The lower AUC values indicate better overall performance across sample sizes of selection methods over random baseline.}
    \label{fig:efficient_auc}
\end{figure}

\subsection{Efficient Prompt Selection}
We use \dataname{} to answer the following question: \emph{How should the values for the different dimensions be chosen to optimize performance, given a fixed inference budget?} This is a practical question whose answer can benefit downstream applications in various real-world scenarios.

Given a set of all possible prompts $C$ and a limited sampling budget $m$, \dataname{} allows us to explore how to efficiently identify prompts that are likely to yield good performance. 
This question has actionable practical implications, as evaluating all possible prompts is computationally prohibitive.

We leverage \dataname{} to simulate different sampling scenarios, focusing on zero-shot settings. For each model, we establish ground truth by finding the prompt $c^* \in C$ that maximizes performance across our complete dataset. We then investigate how different selection methods perform with limited number of samples.

In particular, we explore four strategies for choosing a prompt based on a set of observations: (1) \emph{independent selection:} chooses the best observed value for each dimension, marginalizing all other dimensions; (2) \emph{linear regression:} we train a linear regression on the observed samples which aims to predict accuracy from the set of discrete observed values for each dimension; (3) \emph{maximum observed prompt:} chooses the values for all dimensions according to the best performing prompt in the observed set; and (4) \emph{random baseline:} chooses the values for all dimensions at random. 


Figure~\ref{fig:efficient} shows the accuracy of the different approaches along various data sizes, reporting for each the mean accuracy as well as its standard deviation across 10 random seeds, while Figure~\ref{fig:efficient_auc} shows the area under the graph for each of the the different approaches (See Appendix~\ref{sec:appn:selection_methods_across_all_models} for similar results across all models).

It is evident that different prompt selection approaches can lead to vastly different results. 
Interestingly, choosing the values for the different dimensions in an independent manner achieves performance on par with linear regression, and performs better than choosing the best observed performance. Choosing the best observed prompt becomes reliable with more data, but only after observing tens of millions of samples. 





\begin{figure*}[tb!]
    \centering
  \includegraphics[width=\linewidth]{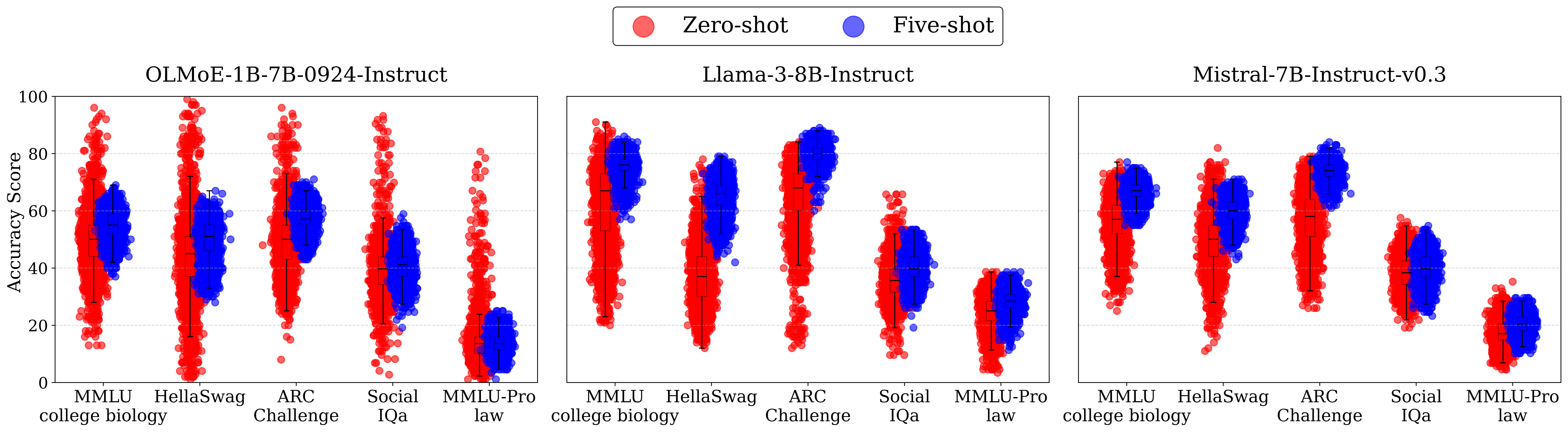}
\caption{\textbf{Few-shot reduces performance variance across evaluation dimensions}. Comparing zero-shot and five-shot on a subset of domains from \dataname{} reveals a narrower spread of accuracy scores. Each point represents the accuracy across 100 instances, demonstrating that the five-shot demonstrations lead to more robust performance.}
  
    \label{fig:zero_vs_few_shot}
\end{figure*}
\subsection{Few-Shot Demonstrations Consistently Reduce Sensitivity}


Figure~\ref{fig:zero_vs_few_shot} depicts the performance of prompts with few-shot demonstrations versus zero-shot prompts. We find that few-shot demonstrations consistently lead to more robust performance (see Appendix ~\ref{sec:performance_few_shot_analysis_all_domains} results across all domains). 

Still, few-shot demonstrations are far from completely mitigating all sensitivity. Even with demonstrations we see a wide range of scores, e.g., above 20\% for all datasets in Figure~\ref{fig:zero_vs_few_shot}. Furthermore, their effect is sometimes minimal, for example, in Social IQa and in the legal domain of MMLU-Pro. 

From a practical perspective, these results suggest that few-shot examples should be added where possible to mitigate the sensitivity of current LLMs.

\begin{figure*}[tb!]
    \centering
  \includegraphics[width=\linewidth]{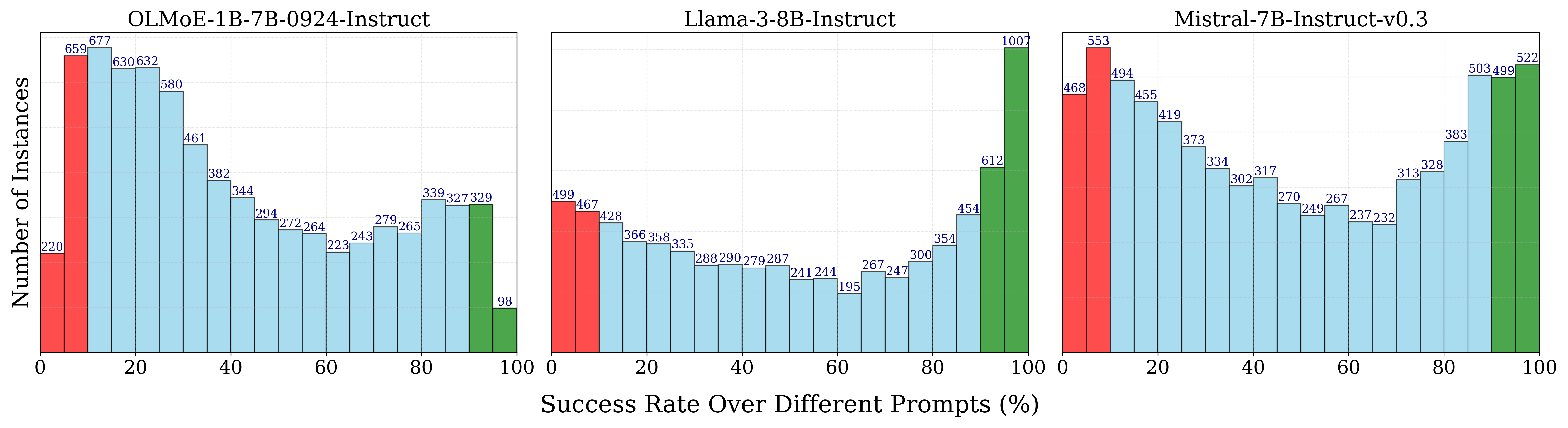}
    \caption{\textbf{Success rate distribution reveals inherent example difficulty patterns.}
    Distribution of success rates by evaluation dimension and model. The x-axis shows the percentage of successful perturbations per instance, while the y-axis shows the instance count in \dataname{}. The distribution reveals examples that are consistently easy or difficult for LLMs across prompt dimensions.}
    \label{fig:model_histograms_grid}
\end{figure*}


\subsection{Some Examples are Consistently Easy or Consistently Hard for Models}
We use \dataname{} to perform an instance-level analysis. Figure~\ref{fig:model_histograms_grid} categorizes each sample according to its \emph{success rate}, which we calculate as the percentage of prompt perturbations for which the model outputs the correct answer, of all the perturbations for that sample. The lower ends of the spectrum, marked in red, count instances for which the model errs on all prompt perturbations, whereas on the higher end of the spectrum are samples for which the models succeds on all prompt perturbations.

These results suggest a novel definition for what constitutes inherently hard instances for models, namely where they fail on all possible prompt perturbations for the same instance. Moreover, on either of these extreme ends, models are in fact \emph{less} sensitive, as they consistently succeeded or err on all prompt perturbations. 




\section{Call for Community Collaboration}
\label{sec:call_community_collaboration}
Our public release of \dataname{} provides a comprehensive dataset and standardized schema for multi-dimensional LLM evaluations.
Yet, the scale and scope of LLM evaluation demands collaborative efforts beyond the capacity of individual research groups.
By establishing \dataname{} as an open and extensible platform, we invite the research community to contribute existing model evaluations, enhancing the collective understanding of prompt sensitivity across diverse models, tasks, and languages. In this manner, we can also improve the efficiency and accessibility of research into LLM evaluation. Instead of rerunning the same evaluation by independent groups unaware of each other's work, a public and standardized repository can provide a reference which developers can browse and enrich.\footnote{\url{https://slab-nlp.github.io/DOVE}}

\paragraph{A Living, Evolving Benchmark}
\dataname{} is designed as a dynamic, evolving benchmark that grows continuously through community contributions and ongoing analyses. Recent updates include evaluations from models such as Llama-3.3-70B, 
incorporated by converting evaluation data collected in a recent study~\cite{lior2025reliableevalrecipestochasticllm}, which we convert to our standardized format. 
The evolving nature of \dataname{} ensures up-to-date insights into model behavior and evaluation robustness.


\paragraph{What to Contribute?}
\dataname{} invites model predictions on existing benchmarks, especially from novel architectures or approaches. Converting public datasets into the \dataname{} format expands coverage to new tasks, domains, and languages, particularly underrepresented or specialized areas. Methodological innovations and suggestions for new evaluation benchmarks are also encouraged.

\paragraph{How to Contribute?}
Contributions can be made through pull requests to our HuggingFace repository or coordinated directly via email for larger or more complex submissions. Researchers seeking to include additional datasets or models in \dataname{} are encouraged to send requests specifying the desired datasets, models, or evaluation parameters clearly. We will properly attribute and reference all contributors.


\section{Related Work}
\label{sec:Related_Work}

Many studies which we have leveraged extensively throughout this work have focused on individual prompt dimensions, examining variations in instruction wording~\cite{mizrahi-etal-2024-state,leidinger-etal-2023-language,sclar2024quantifyinglanguagemodelssensitivity}, answer ordering~\cite{gupta2024changinganswerorderdecrease,Wang2024LLMsMP}, input perplexity~\cite{gonen-etal-2023-demystifying}, few-shot example selection~\cite{reif-schwartz-2024-beyond,lu-etal-2022-fantastically}, and answer enumeration styles~\cite{alzahrani-etal-2024-benchmarks}.
Some works propose metrics for prompt sensitivity,
such as POSIX~\cite{Chatterjee2024POSIXAP}, which measures log-likelihood shifts, and ProSA~\cite{zhuo2024prosaassessingunderstandingprompt}, which uses decoding confidence. Although these methods quantify sensitivity, they do not examine interactions between multiple perturbations, nor do they collect data and make observations at a large scale.

Several recent work have noted that similarly to our findings, few-shot examples help improve performance~\cite{webson-pavlick-2022-prompt,perez2021true}. 
In contrast to these works, we show the effect that few-shot examples have on reducing prompt sensitivity.

Beyond investigating individual factors, several notable frameworks aim to standardize and improve evaluation process. HELM~\cite{liang2023holistic} takes a broad view of LLM performance by creating a taxonomy of a wide range of use cases and evaluation metrics, but was not designed to examine prompt sensitivity. OLMES~\cite{gu2024olmes} establishes detailed protocols for the reproducibility of the evaluation, carefully specifying aspects such as prompt formatting. OLMES demonstrated that standardizing these procedures could lead to more consistent results but may inadvertently harm models which do not perform well on its specific dimension choices.

Although these studies have provided valuable insights, our work is the first to take a holistic view of the problem. This large-scale dataset, encompassing more than \datasetsize{} model predictions, allows us to aggregate across multiple prompt dimensions, noticing practical patterns, and opening the door for many future research directions.

\section{Future Work}
\label{sec:future_work}

\dataname{} provides a foundation for exploring LLM evaluation and sensitivity. The dataset's broad coverage enables flexible partitioning for granular error analysis, targeted evaluations, and investigations of specific dimensions. Future research directions include understanding model biases, improving evaluation methodologies, and refining confidence estimation.
 



    
\paragraph{Task-level sensitivity:} Do some model capabilities have distinct sensitivity patterns? For example, is factual retrieval more fragile than logical reasoning? Do format biases manifest differently across tasks from different domains?
    
\paragraph{Alternative evaluation measures:} Do less common approaches, like perplexity-based evaluation or sensitivity-aware assessments, better mitigate prompt sensitivity in benchmarks?~\cite{gonen-etal-2023-demystifying} Do past prediction data help predict the most effective evaluation method for a new benchmark?~\cite{polo2024sloth,maia2025efficient}

\paragraph{Optimizing evaluation focus:} Given resource constraints, what dimensions are most critical for assessing model performance? Can a predictive framework identify the relative importance of different dimensions?
    
\paragraph{Instance characterization:} What distinguishes consistently answered examples from those with high variability, e.g., as expemlified by the two ends of the spectrum in Figure~\ref{fig:model_histograms_grid}? Do specific linguistic, semantic, or structural features influence susceptibility to example variation?
    
\paragraph{Uncertainty quantification:} How do token-level log probabilities relate to model consistency? Can their distributions help predict or explain model sensitivity better than accuracy scores? Towards that goal \dataname{} also records all model log probabilities.

\paragraph{Future versions of \dataname:}
We plan to expand \dataname{} through both our team's ongoing efforts and community contributions. 
To facilitate community contributions to \dataname{}, we will release tools and documentation to expand coverage across domains, languages, and  tasks. 
We particularly welcome contributions that extend coverage to specialized domains and tasks.
    

\section{Conclusions}
\label{sec:conclusions}

We introduced \dataname{}, a large-scale dataset of \datasetsize{} model predictions across prompt dimensions. Our analysis revealed prompt sensitivity remains a significant challenge, with performance varying over 10\% across different prompt variations. Key findings showed dimension-wise tuning outperforms entire-prompt optimization, few-shot demonstrations reduce but do not eliminate sensitivity, and certain examples remain challenging across all prompt variations. The public release of \dataname{} aims to democratize evaluation research and enable development of robust protocols for assessing LLM capabilities.


\section{Limitations}
\label{sec:limitations}


While \dataname{} provides valuable insights into LLM evaluation, several limitations should be acknowledged. Our focus on multiple-choice questions, while enabling controlled study of prompt variations, does not capture the full complexity of open-ended generation tasks. However, multiple-choice questions remain a fundamental benchmark in the field, with most models reporting results on such tasks. Though we explore various prompt dimensions including paraphrasing, enumeration, and ordering based on prior work, the exponential space of possible variations necessitates a selection of dimensions and values. We plan to systematically expand these dimensions based on analyses of the current version. Additionally, despite its scale, \dataname{} is currently constrained in terms of model diversity and language coverage, and we plan to expand to additional languages and domains in the next version. Large-scale prompt variations computational costs constrain update frequency. We welcome community contributions to expand the \dataname{} scope.

\section{Acknowledgements}
\label{sec:acknowledgements}
This research was conducted in collaboration with the Hebrew University of Jerusalem and IBM Research. The work was supported by the IBM-HUJI Research collaboration. This research was also supported by the Ministry of Innovation, Science \& Technology, Israel (Grant No. 0008239). We gratefully acknowledge the support of both institutions in facilitating this research. We thank Oyvind Tafjord and Jiangjiang Yang for insightful discussions and thoughtful insights.

\bibliography{acl_latex}

\appendix



\onecolumn
\section{Prompt Dimensions Values}
\label{sec:appendix_templates}

\begin{table*}
    \renewcommand{\arraystretch}{1.2} 
    \setlength{\tabcolsep}{6pt} 
    \begin{tabular}{p{4cm} p{8cm}} 
        \toprule
        \textbf{Dimension} & \textbf{Possible Values} \\
        \midrule
        \multirow{6}{*}{\textbf{Enumerator} } 
            & ``A, B, C, D..'' (Capitals) \\
            & ``a, b, c, d...'' (Lowercase) \\
            & ``1, 2, 3, 4...'' (Numbers) \\
            & ``I, II, III, IV...'' (Roman numerals) \\
            & ``\textdollar ! @ \# \% \textasciicircum{}...'' (Keyboard symbols) \\
            & ``$\alpha$, $\beta$, $\gamma$, $\delta$'' (Greek letters) \\
        \midrule
        \multirow{7}{*}{\textbf{Choice Separator}  } 
            & \texttt{"\textbackslash s"} (Space) \\
            & \texttt{"\textbackslash n"} (Newline) \\
            & \texttt{", "} \\
            & \texttt{"; "} \\
            & \texttt{" | "} \\
            & \texttt{" OR "} \\
            & \texttt{" or "} \\
        \midrule
        \multirow{7}{*}{\textbf{Choices Order }  } 
            & Keep original order \\
            & Sort by length (ascending) \\
            & Sort by length (descending) \\
            & Sort alphabetically (ascending) \\
            & Sort alphabetically (descending) \\
            & Force correct choice at first index \\
            & Force correct choice at last index \\
        \bottomrule
    \end{tabular}
    \caption{\textbf{Prompt Formatting Dimensions}. 
\textit{Prompt Formatting Dimensions.} We systematically vary the dimensions when creating prompts. 
\textit{Enumerator} controls how answer options are labeled, 
\textit{Choice Separator} determines how answer options are delimited, 
and \textit{Choices Order} rearranges (or fixes) the position of the correct choice position.}
\label{tab:prompt-formatting-dimensions}
\end{table*}

We present the complete set of prompt dimensions used to build \dataname{}, including enumerators, choice separators, and choice ordering options (see Table~\ref{tab:prompt-formatting-dimensions}). Below, we provide the full collection of instruction phrasings that used in \dataname{}.
\\
\label{sec:instruction_phrasing_options}
\begingroup
\sloppy
\setstretch{0.9}
\lstset{
    basicstyle=\ttfamily\footnotesize,
    aboveskip=0pt, belowskip=0pt,
    showstringspaces=false,
    breaklines=true
}

\begin{lstlisting}
The following are multiple choice questions (with answers) about {topic}.
{question}
{choices}
Answer:
\end{lstlisting}
\vspace{7pt}

\begin{lstlisting}
The following are multiple choice questions (with answers).
{question}
{choices}
Answer:
\end{lstlisting}
\vspace{7pt}
\begin{lstlisting}

Question: {question}
{choices}
Answer:
    \end{lstlisting}
    \vspace{5pt}
    \begin{lstlisting}
The following are multiple choice questions (with answers).

Question: {question}

{choices}
Answer:
    \end{lstlisting}
    \vspace{5pt}
    \begin{lstlisting}
Question: {question}

Choices: {choices}
Answer:
    \end{lstlisting}
    \vspace{5pt}

    \begin{lstlisting}
Topic: {topic}
Question: [question] Choices: [choices] Answer: [answer]
Question: {question} Choices: {choices} Answer:
    \end{lstlisting}
    \vspace{5pt}

    \begin{lstlisting}
Question: [question] Choices: [choices] Answer: [answer]
Question: {question} Choices: {choices} Answer:
    \end{lstlisting}
    \vspace{5pt}
    \begin{lstlisting}
Please answer the following question:
{question}
{choices}
Answer:
    \end{lstlisting}
    \vspace{5pt}

    \begin{lstlisting}
Please address the following question:
{question}
{choices}
Answer:
    \end{lstlisting}
    \vspace{5pt}

    \begin{lstlisting}
Could you provide a response to the following question:
{question}
{choices}
Answer:
    \end{lstlisting}
    \vspace{5pt}
    \begin{lstlisting}
Here are some multiple choice questions along with their answers about {topic}.

Question: {question}
Choices: {choices}
Correct Answer:
    \end{lstlisting}
    \vspace{5pt}

    \begin{lstlisting}
Below are multiple-choice questions related to {topic}, each followed by their respective answers.

Question: {question}
Choices: {choices}
Correct Answer:
    \end{lstlisting}
    \vspace{5pt}

    \begin{lstlisting}
Below are multiple-choice questions related to {topic}. Please provide the correct answer for each question.

Question: {question}
Choices: {choices}
Answer:
    \end{lstlisting}

\endgroup




\twocolumn

\section{Extended Results}
\label{sec:appendix_more_results}

    

    
    

\subsection{Performance Analysis Across All Domains}
\label{sec:performance_analysis_all_domains}

\begin{figure*}[tb!]
    \centering
    \includegraphics[width=\linewidth]{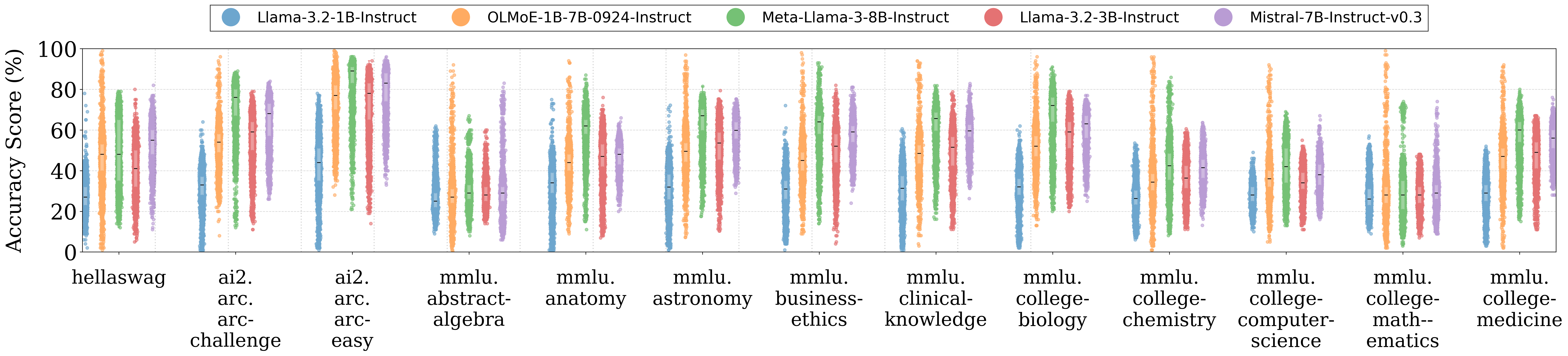}
    \includegraphics[width=1\linewidth]{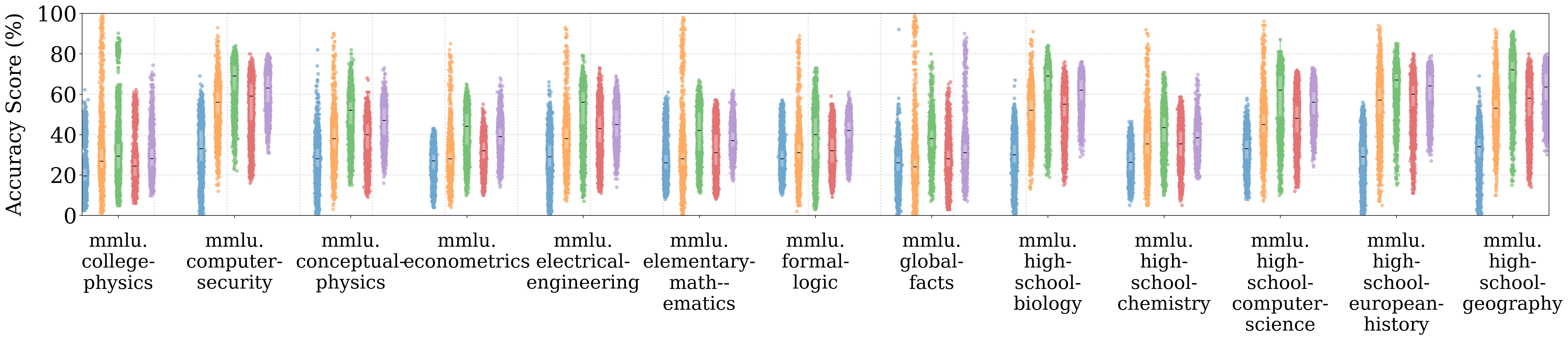}

    \includegraphics[width=1\linewidth]{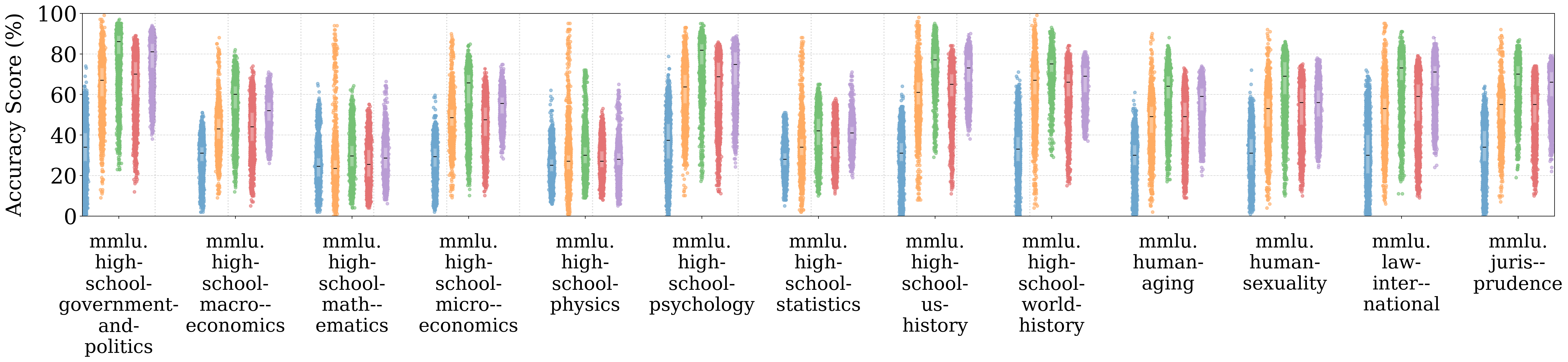}

    \includegraphics[width=1\linewidth]{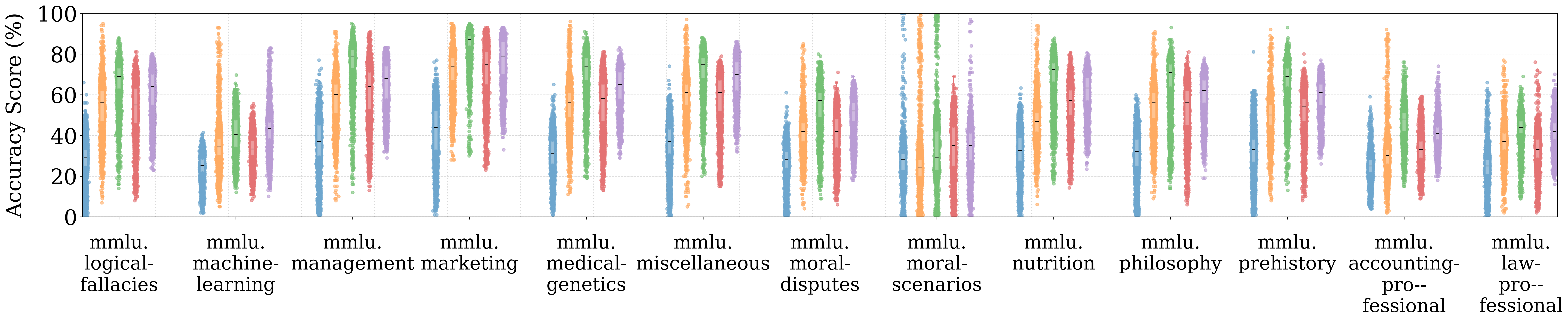}

    \includegraphics[width=1\linewidth]{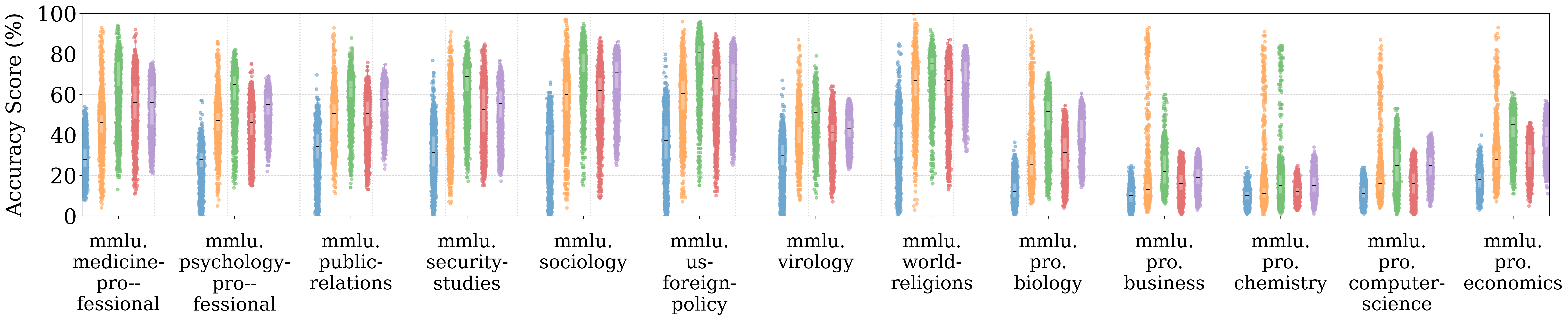}

    \includegraphics[width=1\linewidth]{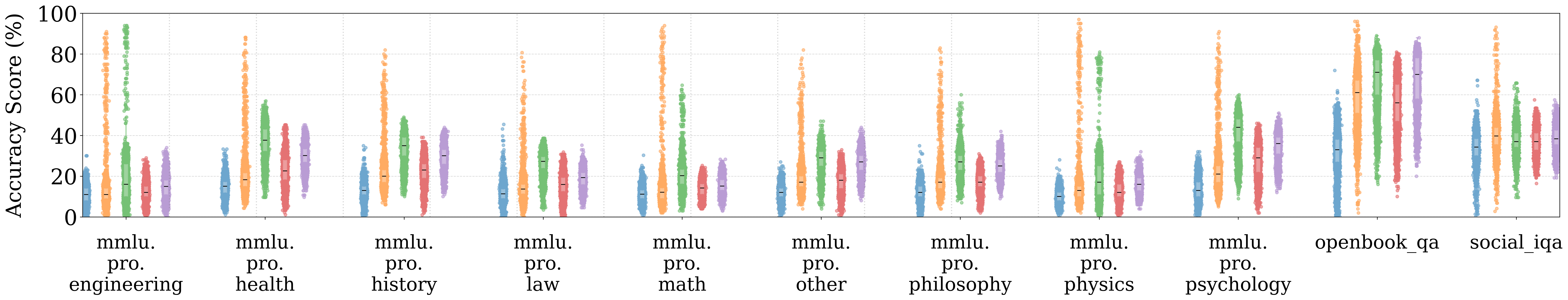}
    
\caption{Performance variations across all evaluation domains (shown in standard deviations).}
\label{fig:performance_analysis_all_domains}

\end{figure*}

Figure~\ref{fig:performance_analysis_all_domains} reveals consistent patterns in prompt sensitivity across our evaluation domains.
\subsection{Analysis of Few-Shot Impact Across All Domains}
\begin{figure*}[tb!]
    \centering
    \includegraphics[width=\linewidth]{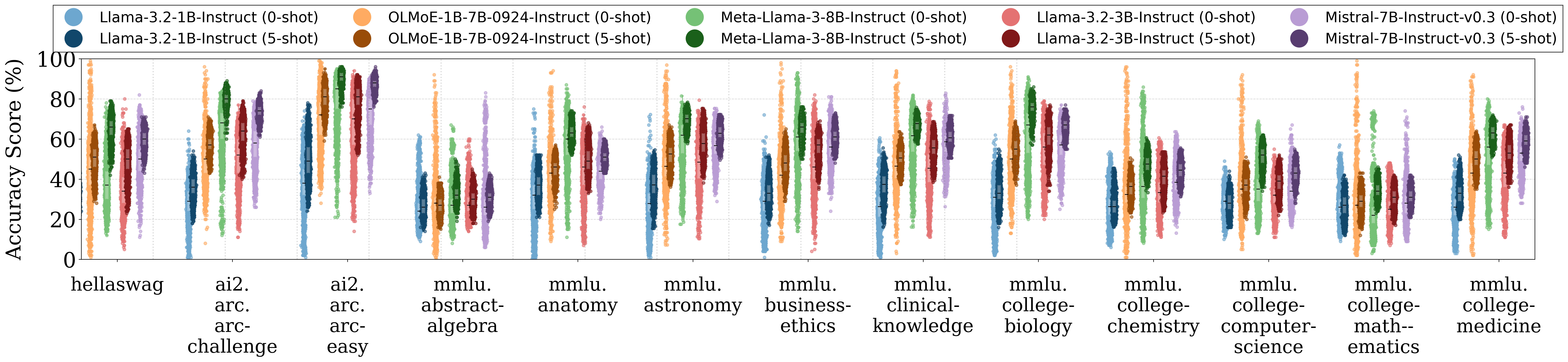}
    \includegraphics[width=1\linewidth]{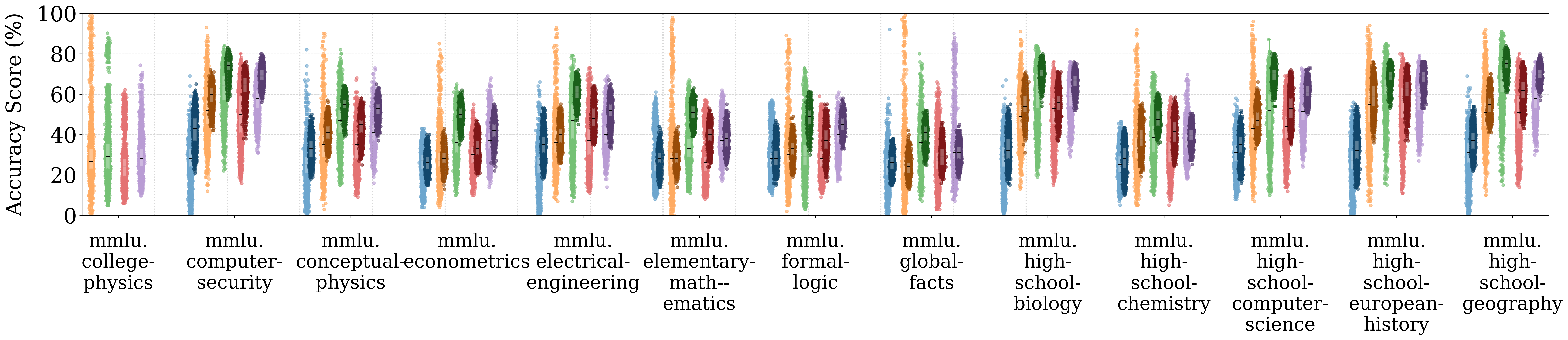}

    \includegraphics[width=1\linewidth]{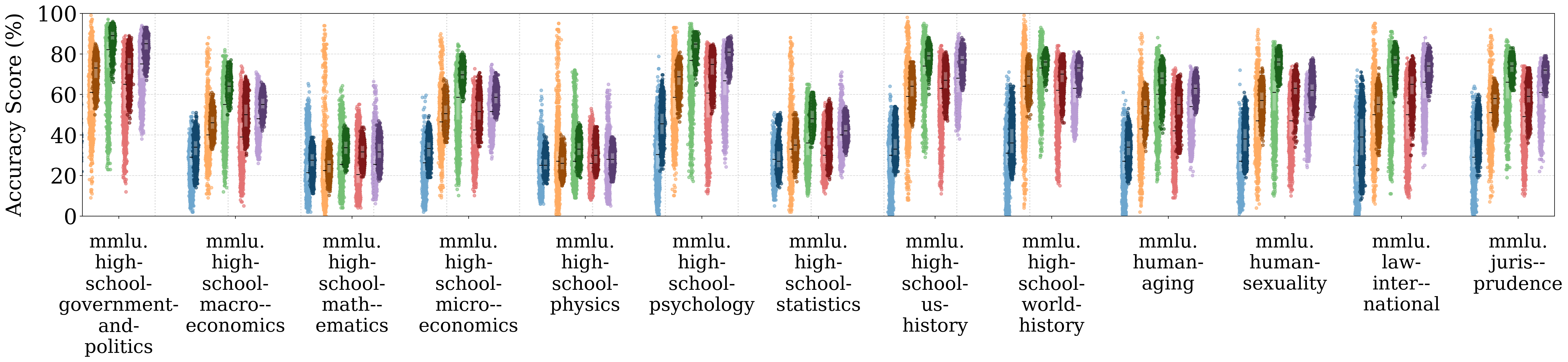}

    \includegraphics[width=1\linewidth]{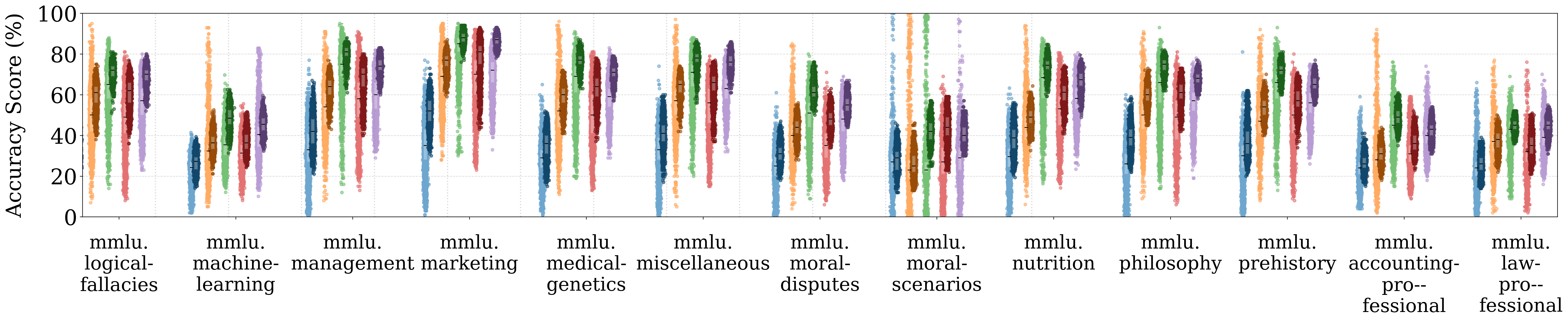}

    \includegraphics[width=1\linewidth]{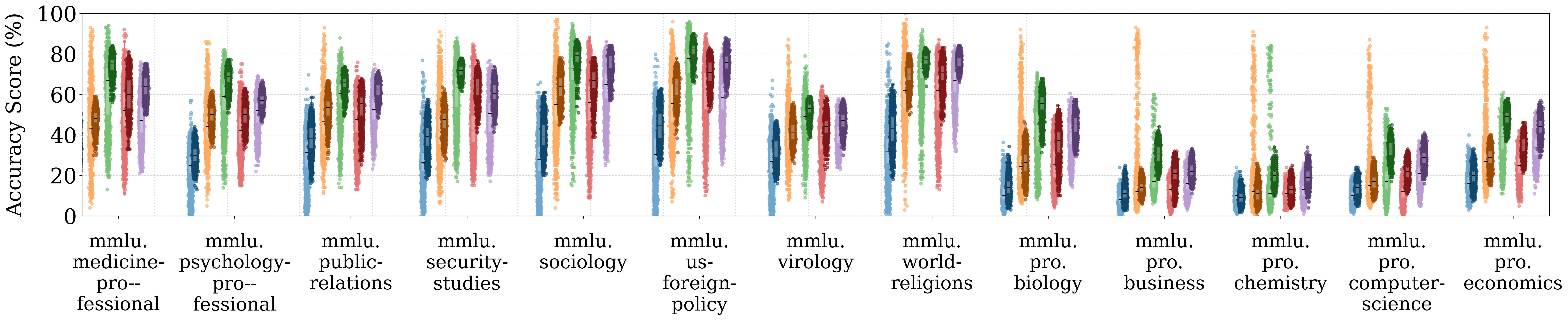}
    \includegraphics[width=1\linewidth]{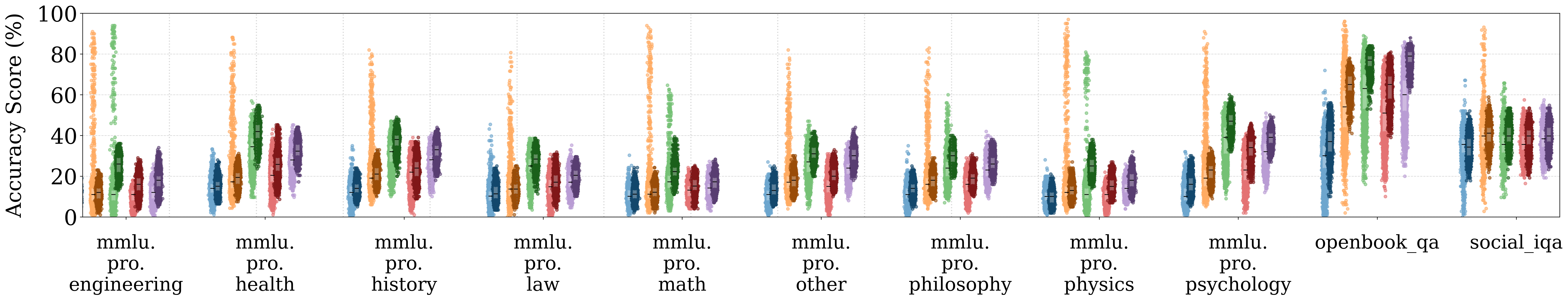}
    
\caption{Few-shot versus zero-shot performance across all domains. Extended analysis showing consistent reduction in sensitivity with few-shot demonstrations}

    \label{fig:few_shot_performance_analysis_all_domains}
\end{figure*}

The impact of few-shot demonstrations on reducing prompt sensitivity becomes evident across domains, as illustrated in Figure~\ref{fig:few_shot_performance_analysis_all_domains}.

\label{sec:performance_few_shot_analysis_all_domains}

\subsection{Divergence Across All Domains}
\begin{figure*}
    \centering
  \includegraphics[width=\linewidth]{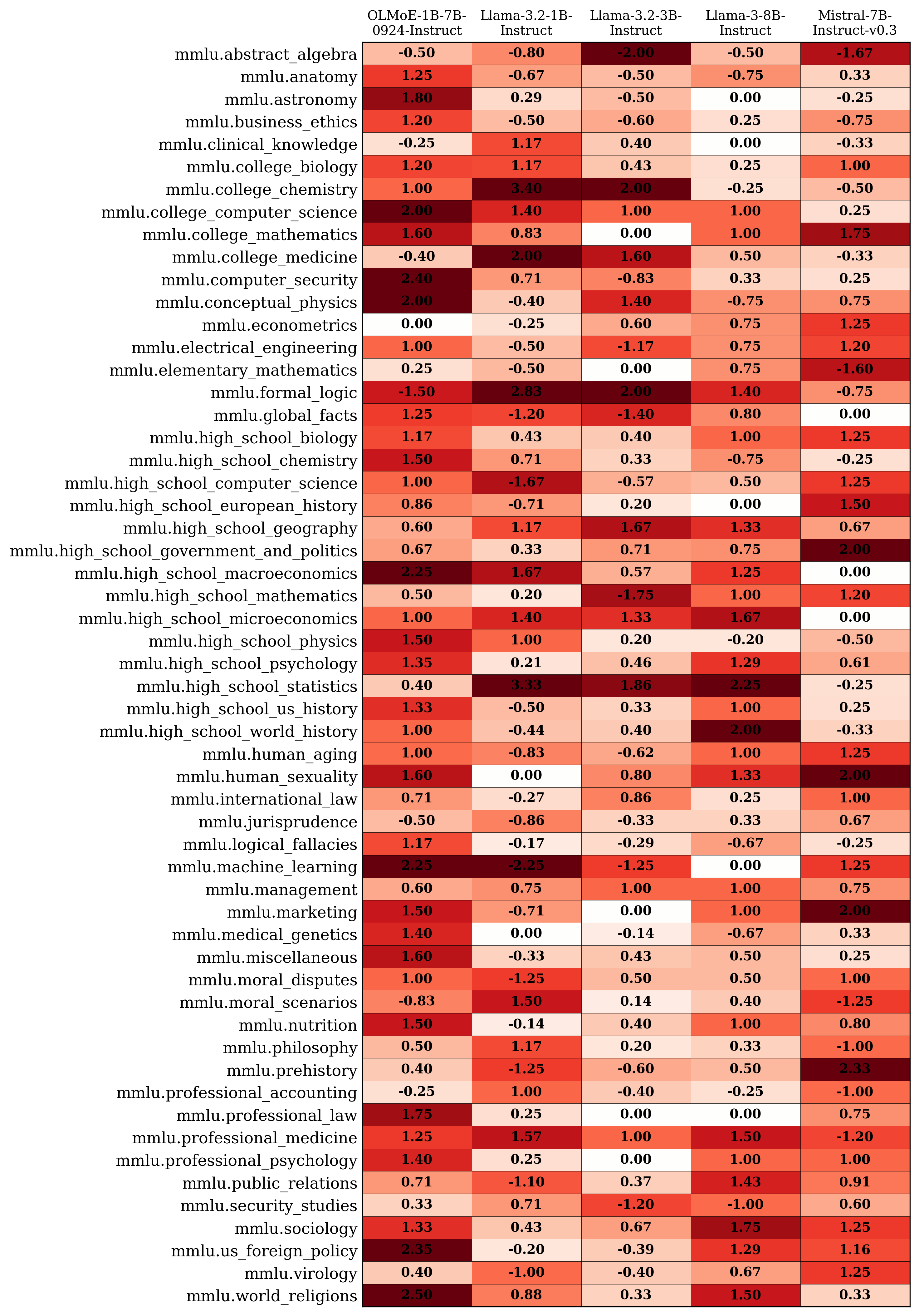}
  \caption{Model performance variations across different prompts perturbation (shown in standard deviations).}
  \label{fig:appendix_divergence_plot}
 \end{figure*}

Figure~\ref{fig:appendix_divergence_plot} highlight the
variations across all datasets.

\label{sec:divergence_across_all_domains}
 


\subsection{Selection Methods Across All Models}
\begin{figure*}[tb!]
    \centering
    \includegraphics[width=\linewidth]{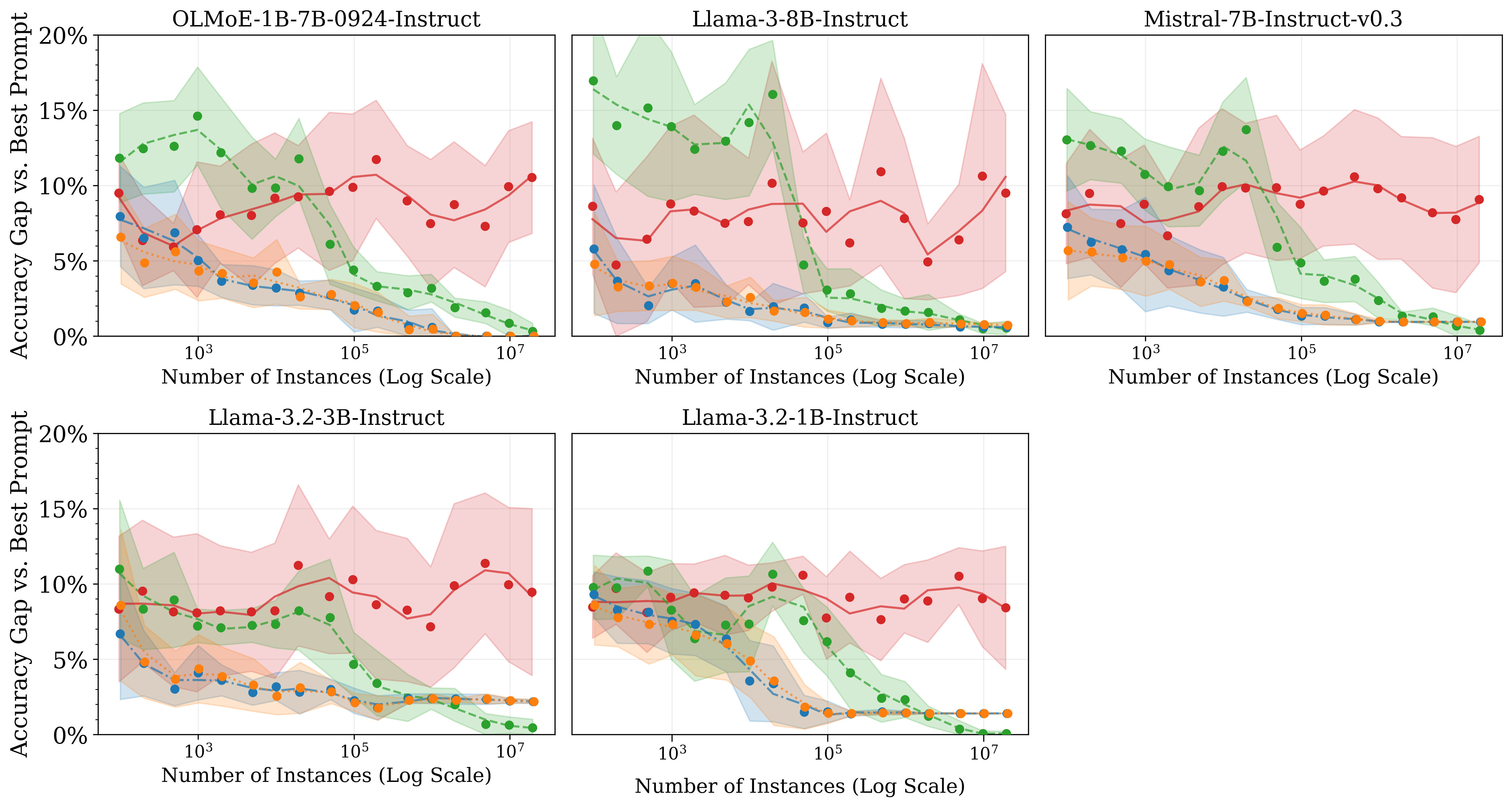}
    
    \includegraphics[width=1\linewidth]{figures/prompt_selection_legend.png}
    \caption{Efficient prompt selection across all models}
    \label{fig:append_comparison_methods}
\end{figure*}

\begin{figure}[H]
    \centering
    \includegraphics[width=\linewidth]{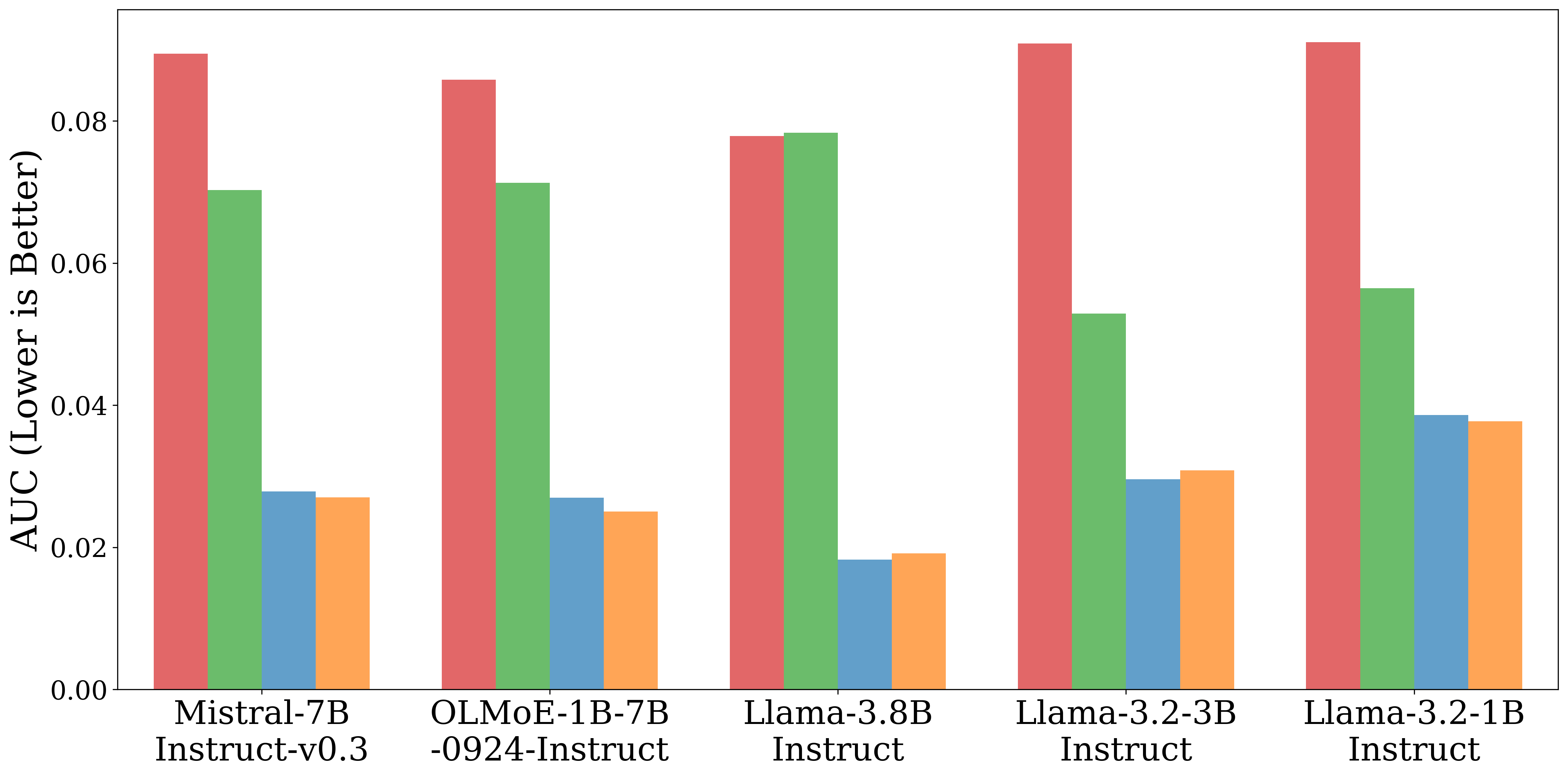}
     
    \includegraphics[width=0.8\linewidth]{figures/combined_figure_auc_legend.png}

    \caption{AUC comparison of prompt selection methods across all models} 
    
    \label{fig:optimized_selection}
\end{figure}

Our comparison of prompt selection methods covers both AUC analysis and success rate distributions. Results are shown in Figure~\ref{fig:append_comparison_methods} and
Figure~\ref{fig:optimized_selection}.
\label{sec:appn:selection_methods_across_all_models}

\begin{figure*}[tb!]
    \centering
  \includegraphics[width=\linewidth]{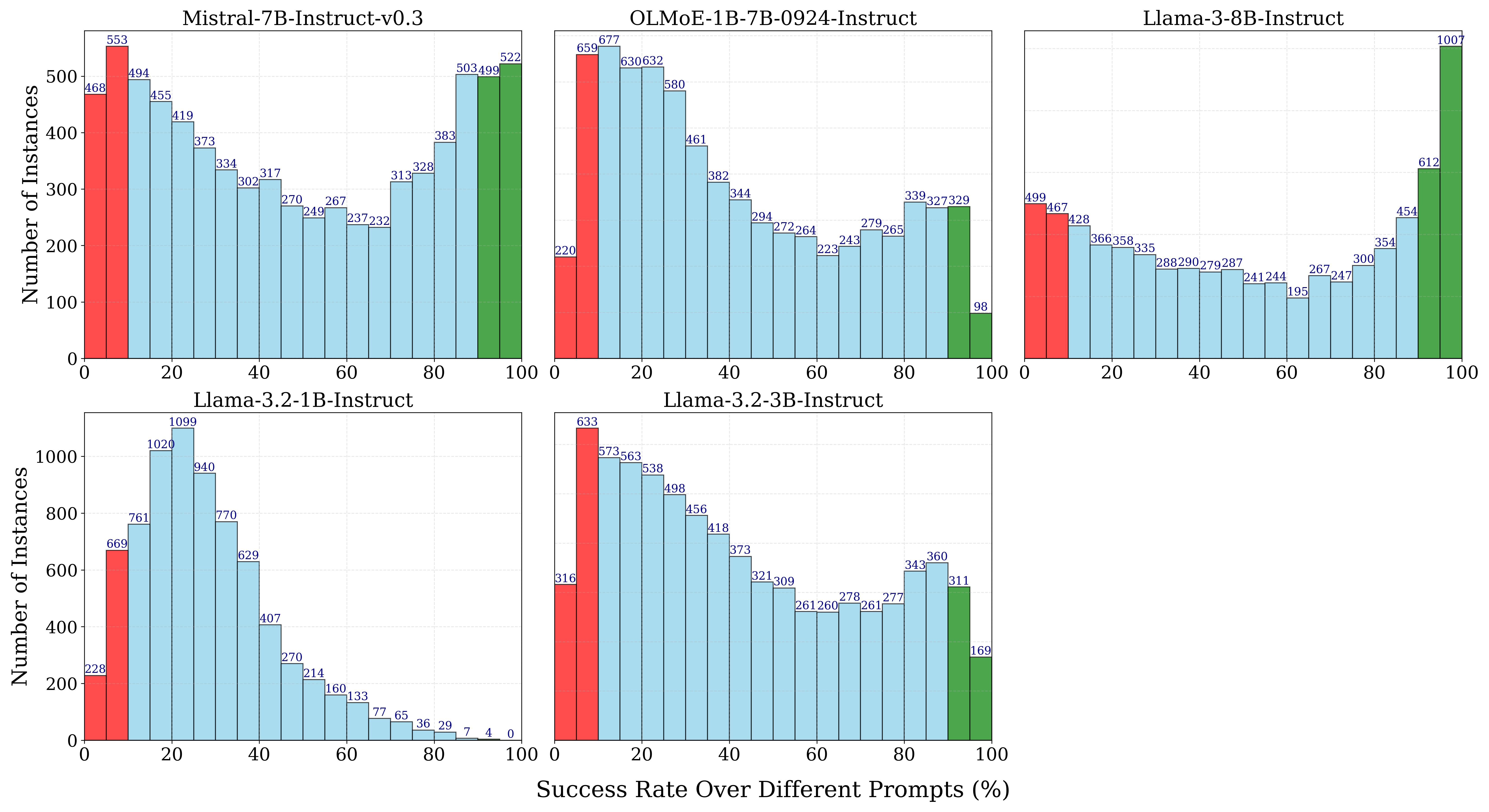}
\caption{Success rate distribution reveals
inherent example difficulty patterns}
    \label{fig:appendix_histograms}
\end{figure*}

\subsection{Examples are Consistently Easy or
Hard Across All Models}
Task difficulty follows consistent patterns across different models, with success rate distributions mapped in  Figure~\ref{fig:appendix_histograms}.
\label{sec:success_rate_distributions_across_all_models}

\section{Dataset Scheme}
Table~\ref{table:dataset_format} details the components and structure of our dataset, providing descriptions and example values for each field.

\label{sec:appendix_dataset_scheme}

\begin{table*}[htb!]
\centering

\resizebox{\textwidth}{!}{
\begin{tabular}{p{1.7cm} p{3.5cm} p{6.5cm} p{5cm}}
\hline
\textbf{Component}    & \textbf{Field}                  & \textbf{Description}                                      & \textbf{Example Values} \\ \hline
\textbf{ID}          & Evaluation ID                   & Unique identifier for the evaluation run                 & \texttt{f8544...2240} \\ \hline
\textbf{Model}       & Name                             & Model identifier and version                           & \texttt{Mistral-7B-Instruct-v0.3} \\
                     & Configuration                         & Architecture, Size, Context window, Instruction tuning & \texttt{transformer, 7B, 32768, True} \\
                     & Quantization                     & Bit precision and method settings for model inference  & \texttt{float16, none} \\
                     & Generation Args                  & Generation control settings                            & \texttt{temperature:null, top\_p:null, top\_k: -1} \\ \hline
\textbf{Instance}    & Task Type                        & Type of evaluation task                                  & \texttt{classification, generation} \\
                     & Raw Input                        & Original input from the dataset (before formatting)      & \texttt{"What size of cannula would you use..."} \\
                     & Sample Identifier                & Dataset source details, including split and index       & \texttt{mmlu.clinical\_knowledge, test, 487} \\
                     & Language                         & Language of the input                                   & \texttt{en, fr, ar, zh} \\
                     & Tokens Logprobs                  & Log probability of prompt tokens                        & \texttt{[{token\_index:153, logprob:-0.96, rank:1, decoded\_token:"Question"}, ...]} \\
                     & Classification Fields            & Classification details: question, choices, answer       & \texttt{question, choices, gt} \\ \hline
 
\textbf{Prompt} & Prompt Class &  Type of formatting requirements & MultipleChoice  \\ \cdashline{1-4}
Dimensions
&  Instruction Phrasing  & Template text with placeholders & "Below are multiple-choice questions..." \\
& Separator & Character(s) used to separate multiple-choice options & \texttt{"\textbackslash s", "\textbackslash n", ", ", " | ", " OR ", " or "} \\
& Enumerator & Style of enumeration for multiple-choice options & \texttt{"ABCD", "abcd", "1234", "I,II,III,IV", "!@\#\$", "$\alpha\beta\gamma\delta$"} \\
&  Choices Order  & Method for ordering answer choices & \texttt{"original order, by length, alphabetical, correct first/last"} \\
&  Shots & Number of examples included in the prompt & \texttt{"zero, two, five"} \\
&  Demonstrations  & Array of example question-answer pairs & \texttt{"question, choices, answers"} \\
\hline

\textbf{Output}      & Response                         & Model's full response to the prompt                   & \texttt{"The size depends on a number of factors..."} \\
                     & Tokens Logprobs                  & Log probabilities for generated tokens                & \texttt{[{token\_index:1183, logprob:-2.73, rank:4, decoded\_token:"The"}, ...]} \\
                     & Cumulative Logprob               & Log probability of the entire generated sequence      & \texttt{-49.28} \\ \hline
\textbf{Evaluation}  & Ground Truth                     & The correct answer for the given instance             & \texttt{"IV. 18 gauge."} \\
                     & Evaluation Method                & Method used to evaluate the model’s response          & \texttt{label\_only\_match, content\_similarity} \\
                     & Score                            & Binary score indicating correctness                   & \texttt{1} \\ \hline
\end{tabular}
} 
\caption{Dataset Schema Components, Descriptions, and Example Values}
\label{table:dataset_format}
\end{table*}


\end{document}